\documentclass{article}

 \usepackage[dblblindworkshop, final]{neurips_2025}
\workshoptitle{DL4CODE}

\usepackage[utf8]{inputenc} 
\usepackage[T1]{fontenc}    
\usepackage{hyperref}       
\usepackage{url}            
\usepackage{booktabs}       
\usepackage{amsfonts}       
\usepackage{nicefrac}       
\usepackage{microtype}      
\usepackage{xcolor}         
\usepackage{amsmath}
\usepackage{multirow}
\usepackage{subcaption}

\DeclareRobustCommand{\mytexttt}[1]{%
  {\texttt{\hyphenchar\font=`\- #1}}%
}

\newcommand{\model}{\href{https://huggingface.co/collections/jinaai/jina-code-embeddings-68b0fbfbb0d639e515f82acd}{\texttt{jina-code-embeddings}}}
\newcommand{\modelS}{\href{https://huggingface.co/jinaai/jina-code-embeddings-0.5b}{\texttt{jina-code-embeddings-0.5b}}}

\newcommand{\modelLshort}{\href{https://huggingface.co/jinaai/jina-code-embeddings-1.5b}{\texttt{1.5b}}}

\title{Efficient Code Embeddings from Code Generation Models}

\author{%
  Daria Kryvosheieva\textsuperscript{1,2}\thanks{Work done during internship at Jina AI.} \quad Saba Sturua\textsuperscript{2} \quad Michael G\"{u}nther\textsuperscript{2} \\
  \textbf{Han Xiao\textsuperscript{2}} \\
  \\
  \textsuperscript{1}Massachusetts Institute of Technology\quad
  \textsuperscript{2}Jina AI GmbH \\
  Prinzessinnenstraße 19, 10969, Berlin, Germany \\
  \texttt{research@jina.ai} \\
}
\begin{document}

\maketitle

\begin{abstract}
  \model{} is a novel code embedding model suite designed to retrieve code from natural language queries, perform technical question-answering, and identify semantically similar code snippets across programming languages. It makes innovative use of an autoregressive backbone pre-trained on both text and code, generating embeddings via last-token pooling. We outline the training recipe and demonstrate state-of-the-art performance despite the relatively small size of the models, validating this approach to code embedding model construction.
\end{abstract}

\section{Introduction}

The rapid adoption of AI-powered development environments like Cursor and Claude Code has transformed software engineering, with code embedding models serving as a critical foundation for retrieval and context engineering of these systems.

Code embedding models, starting with \citet{Gu2018}, have evolved into a well-established subfield with dedicated benchmarks~\citep{Husain2019,coir2025} and leaderboards~\citep{mmteb2025}. While code generation models like Codex~\citep{codex2021} can directly synthesize code from natural language prompts, practical code generation requires contextual understanding of existing codebases, API usage patterns, and integration requirements. This naturally positions code generation systems as retrieval-augmented generation (RAG) architectures~\citep{lewis2020}, where embedding models serve as the critical retrieval component.

Despite their importance, current code embedding models face a fundamental training data limitation. Supervised training typically relies on aligned data such as inline comments, documentation strings, and pedagogical examples from technical documentation—sources that provide insufficient semantic grounding for complex real-world development scenarios. In contrast, the abundant unaligned code and natural language documentation used to train modern LLMs remains largely underutilized for embedding model development.

We address this gap by introducing two high-quality code embedding models: \modelS{} and \modelLshort, with 494 million and 1.54 billion parameters, respectively. Our approach implements several key innovations: First, we leverage dedicated pre-trained code generation LLMs as backbones, adapting them specifically for embedding generation. Second, through comprehensive task analysis across functional areas of code embedding applications, we develop targeted training strategies that optimize performance for each use case. The resulting models achieve significant improvements over previous models of comparable size, with benchmark performance competitive with much larger alternatives.

\section{Related Work}

General-purpose semantic embedding models can function as code embedding models, and if large enough and trained on sufficient relevant data, can even perform comparably with specialized models. For example, among recent general-purpose text embedding models, Qwen3 Embedding \citep{zhang-2025} and Gemini Embedding \citep{lee-2025-gemini} perform well on code tasks. However, general-purpose models that support all kinds of texts are very large and expensive to train and deploy.

Many code embedding models are based on variants of the BERT model~\citep{Devlin2019}, like CodeBERT~\citep{Feng2020} and \mytexttt{jina-embeddings-v2-base-code}~\citep{jina2024a}. However, this approach requires specialized datasets and struggles with adequate data sources.

An alternative approach is to take pre-trained general-purpose embedding models and adapt them to code, leveraging their existing language knowledge to compensate for shortcomings in aligned natural language and code data. \citet{Sturua2024} use a LoRA adapter~\citep{Hu2022} to specialize the general-purpose \mytexttt{jina-embeddings-v3} model for code retrieval.

Recent developments have demonstrated that autoregressive decoder architectures can be adapted to generate high-quality embeddings~\citep{lee-2025-nv}. This class of model generates tokens sequentially as a function of the preceding tokens and can be trained using standard denoising objectives and copious quantities of readily available natural language text. They can then be adapted, after pre-training, to produce high-quality embeddings by pooling the token embedding outputs of the last hidden layer and fine-tuning the model. This is the approach used in the Qwen3 embedding model family~\citep{zhang-2025}, and in \mytexttt{jina-embeddings-v4}~\citep{guenther-2025}.

Many different pooling methods have been explored for embedding model generation.
\citet{reimers2019} found \emph{mean pooling} to perform better than \emph{max pooling} or \emph{CLS pooling}\footnote{CLS pooling uses the output embedding of a pre-pended \mytexttt{[CLS]} token as the final embedding vector.} for encoder-based transformer models.
For decoder-only models, embeddings are most commonly generated via last-token pooling~\cite{wang-etal-2024-improving-text}. 
Recently, \citet{lee-2025-nv} have proposed a novel pooling scheme they call \emph{latent attention pooling}, inspired by the transformer architecture, which has trainable weights. They report significant improvements in general embedding performance using this method.

Instruction-tuning for specific domains and task types generally yields improved performance~\citep{su-etal-2023-one}. Adding instructions to each text before generating an embedding trains the network to produce embeddings that reflect that instruction. The Qwen3 model~\citep{zhang-2025}, which implements instruction-tuning, provides for user-generated instructions, making it difficult to optimize performance and leading to uncertainty about how the model will behave.

\section{Model Architecture and Task Prefixes}

\modelS{} and \modelLshort{} employ an autoregressive decoder backbone architecture, i.e., a model which generates tokens sequentially as a function of the preceding tokens. They build on the \mytexttt{Qwen2.5-Coder-0.5B} and \mytexttt{Qwen2.5-Coder-1.5B} backbones \citep{hui-2024}, which are both very compact LLMs. 

The final hidden layer of the LLMs is transformed into an embedding via last-token pooling. We found, after some experimentation, that last-token pooling gave us better performance than mean pooling or latent attention pooling, as documented in Appendix~\ref{ablation}. CLS pooling was not tested, but is generally not favored for decoder-only architectures.

We analyzed downstream code embedding tasks and divided them into five categories: Natural language to code retrieval (\textbf{NL2Code}), technical question answering (\textbf{TechQA}), code-to-code retrieval (\textbf{Code2Code}), code to natural language retrieval (\textbf{Code2NL}), and code to completion retrieval (\textbf{Code2Completion}).

For each of these tasks, we created an instruction string in English that prefixes the text passed to the model, listed in Table~\ref{tab:prefixes}. Different instruction strings are used for queries and for documents.

\begin{table}[t]
\centering

\caption{Task categories and their corresponding instruction prefixes.}
\label{tab:prefixes}

\small
\begin{tabular}{l|l|l}
\toprule
\textbf{Task type} & \textbf{Query prefix} & \textbf{Document prefix} \\
\midrule
\multirow{2}{*}{NL2Code} & \texttt{"Find the most relevant code snippet} & \multirow{2}{*}{\texttt{"Candidate code snippet:\textbackslash n"}} \\
 & \texttt{given the following query:\textbackslash n"} & \\
\hline
\multirow{2}{*}{TechQA} & \texttt{"Find the most relevant answer} & \multirow{2}{*}{\texttt{"Candidate answer:\textbackslash n"}} \\
 & \texttt{given the following question:\textbackslash n"} & \\
\hline
\multirow{2}{*}{Code2Code} & \texttt{"Find an equivalent code snippet} & \multirow{2}{*}{\texttt{"Candidate code snippet:\textbackslash n"}} \\
 & \texttt{given the following code snippet:\textbackslash n"} & \\
\hline
\multirow{2}{*}{Code2NL} & \texttt{"Find the most relevant comment} & \multirow{2}{*}{\texttt{"Candidate comment:\textbackslash n"}} \\
 & \texttt{given the following code snippet:\textbackslash n"} & \\
\hline
\multirow{3}{*}{Code2Completion} & \texttt{"Find the most relevant completion} & \multirow{3}{*}{\texttt{"Candidate completion:\textbackslash n"}} \\
 & \texttt{given the following start of code } & \\
 & \texttt{snippet:\textbackslash n"} & \\
\bottomrule
\end{tabular}
\end{table}

\section{Training}\label{sec:training}

We initialized the model with weights of the pre-trained backbone \texttt{Qwen2.5-Coder-0.5B} and then applied further training with a contrastive objective using the InfoNCE loss function~\citep{oord-2019}.  Pairs of inputs are classed as related or unrelated, and the model learns to embed related items closely together and unrelated items further apart.

We use Matryoshka representation learning~\citep{kusupati-2022} during training to produce truncatable embeddings, so users can make flexible trade-offs between precision and resource usage.

\subsection{Training Data}\label{sec:training-data}

The training data consists of query-document pairs for a variety of code retrieval tasks, largely using docstrings, comments, commit messages, problem statements, and internet forum questions as queries, alongside matching code snippets, diffs, or answers as documents. These pairs have been collected from various sources, including the training splits of MTEB code tasks, the non-MTEB code retrieval dataset CoSQA+, and public datasets created for non-retrieval purposes. We also used GPT-4o~\citep{openai-2024} to synthetically generate datasets when available data is scarce.

Details of the training datasets and synthetic data generation are outlined in Appendix~\ref{datasets}.

\subsection{Procedure}\label{sec:procedure}

In each training step, we sample a batch $B = (q_1, d_1), ..., (q_n,d_n)$ of $n$ query-document text pairs. We generate normalized embeddings for all texts in the selected pairs. We then construct a matrix of similarity values $S_{\text{dense}}(B)$ by calculating the cosine similarity of all combinations of embeddings $\mathbf{q}_i$ and $\mathbf{d}_j$ in $B$. We train by taking the training embedding pairs $(\mathbf{q}_i, \mathbf{d}_i)$ as similar, and all other combinations of $(\mathbf{q}_i, \mathbf{d}_j), i \ne j$ in each batch as dissimilar.

Then, we apply the contrastive InfoNCE loss function $\mathcal{L}_{\text{NCE}}$ \citep{oord-2019} on the resulting matrix of similarity scores.
\begin{equation}
    \mathcal{L}_{\text{NCE}}(S(B), \tau) := -\sum_{i, j=0}^{n} \ln{\sigma(S(B), \tau, i, j)}
    \hspace{4mm}
    \text{where~~}\sigma(S, \tau, i, j) := \frac{e^{S_{i,j}/\tau}}{\sum_{k=0}^{n}e^{S_{i,k}/\tau}}
\end{equation}

where $\tau$ is the temperature and $n$ is the batch size, which increases the weight of small differences in similarity scores in calculating the loss. 
During training, we maintain constant hyperparameters: $\tau = 0.05$, $n = 512$ for the 0.5B parameter model and $n = 256$ for the 1.5B parameter one, and sequence length is 512.

Each new batch $B$ was sampled from one of the training datasets according to pre-configured sampling rates (probabilities), which we treated as hyperparameters. 

Training was for 1500 steps on four 80GB VRAM A100 GPUs. Training the 0.5B parameter model took approximately 8.3 hours, and approximately 12 hours for the 1.5B parameter one. As described in Appendix~\ref{ablation}, we repeated training under three conditions to determine the best pooling method.

\section{Evaluation}

To assess performance on code retrieval, we evaluate the model on the MTEB-CoIR benchmark \citep{coir2025}, which consists of 10 tasks spanning text-to-code, code-to-text, code-to-code, and hybrid code retrieval types. We also evaluate the model on code-related MTEB tasks CodeSearchNetRetrieval, CodeEditSearchRetrieval, HumanEval, MBPP, DS-1000, WikiSQL, and MLQuestions, as well as CosQA+ and our in-house benchmarks. See Appendix~\ref{sec:eval-hparams} for evaluation hyperparameters.
Results are reported in Table \ref{tab:evals}. 

\begin{table}[h!]
\small
\centering

\caption{Evaluation Results on Code Retrieval Tasks}
\label{tab:evals}

\begin{tabular}{l|c|c|c|c|c|c}
\toprule
Benchmark & \mytexttt{JCE-0.5B} & \mytexttt{JCE-1.5B} & \mytexttt{JV4} & \mytexttt{Qw3-0.6B} & \mytexttt{VC3} & \mytexttt{GE-001} \\
\midrule
CoSQA+ & 15.42\% & 16.38\% & 13.29\% & 15.63\% & 13.57\% & \textbf{16.44\%} \\
CoSQA* & 39.25\% & 35.10\% & 29.99\% & 37.75\% & 34.11\% & \textbf{51.94\%} \\
MBPP & 89.01\% & 90.13\% & 89.93\% & 88.29\% & \textbf{94.68\%} & 93.46\% \\
COIR-CSN* & 85.73\% & 86.45\% & 84.03\% & 84.78\% & \textbf{89.35\%} & 81.06\% \\
CSN* & 90.68\% & 91.38\% & 84.84\% & 90.77\% & \textbf{93.92\%} & 91.38\% \\
Doc2Code & 95.98\% & 96.34\% & 91.46\% & 94.77\% & \textbf{97.18\%} & 96.54\% \\
SWE-Bench & 83.00\% & 86.33\% & 81.00\% & 76.12\% & 87.02\% & \textbf{87.40\%} \\
CES* & 83.25\% & \textbf{84.43\%} & 72.75\% & 64.21\% & 80.30\% & 81.69\% \\
CP-FT & 63.00\% & \textbf{65.06\%} & 45.93\% & 38.50\% & 59.24\% & 61.18\% \\
AppsR* & 84.17\% & 86.63\% & 78.32\% & 75.22\% & 93.77\% & \textbf{95.70\%} \\
LeetCode & 57.86\% & 59.075\% & \textbf{59.11\%} & 58.23\% & 58.89\% & 58.40\% \\
CodeChef & 94.03\% & 96.89\% & 87.98\% & 84.29\% & 99.18\% & \textbf{99.55\%} \\
SynText2SQL* & 72.80\% & 73.91\% & \textbf{76.98\%} & 66.91\% & 63.39\% & 59.24\% \\
Spider & 81.65\% & \textbf{82.18\%} & 81.18\% & 81.45\% & 81.99\% & 81.15\% \\
WikiSQL & \textbf{98.31\%} & 98.02\% & 96.06\% & 96.04\% & 95.71\% & 90.94\% \\
CF-MT* & 89.56\% & 89.91\% & 70.07\% & 90.79\% & \textbf{93.47\%} & 64.95\% \\
CF-ST* & 85.73\% & 86.18\% & 85.47\% & 86.43\% & \textbf{90.56\%} & 85.70\% \\
StackOQA* & 91.04\% & 92.37\% & 93.80\% & 89.96\% & \textbf{96.90\%} & 96.02\% \\
DS-1000 & 59.77\% & 62.88\% & 64.11\% & 61.19\% & 69.49\% & \textbf{70.10\%} \\
MLQuestions & \textbf{81.05\%} & 77.46\% & 54.71\% & 60.52\% & 66.87\% & 62.95\% \\
CTOC* & 90.37\% & 92.54\% & 92.23\% & 86.28\% & \textbf{93.49\%} & 92.59\% \\
CTODL* & 41.69\% & 37.319\% & \textbf{46.29\%} & 31.78\% & 38.72\% & 32.84\% \\
CodeChefXLang & 99.70\% & 99.44\% & 92.82\% & 90.94\% & 99.13\% & \textbf{99.79\%} \\
CSN-CC* & 90.41\% & 91.12\% & 83.69\% & \textbf{91.41\%} & 90.09\% & 84.69\% \\
HumanEval & 96.77\% & 98.41\% & 96.74\% & 94.84\% & \textbf{99.77\%} & 98.90\% \\
\midrule
Overall AVG & 78.41\% & 79.04\% & 74.11\% & 73.49\% & \textbf{79.23\%} & 77.38\% \\
MTEB Code AVG & 78.72\% & 78.94\% & 74.87\% & 74.69\% & \textbf{79.84\%} & 76.48\% \\
\bottomrule
\end{tabular}

{\raggedright
\textbf{Models:} \mytexttt{JCE}: \model{}; \mytexttt{JV4}: \mytexttt{jina-embeddings-v4}; \mytexttt{Qw3-0.6B}: \mytexttt{Qwen3-Embedding-0.6B}; \mytexttt{VC3}: \mytexttt{voyage-code-3}; \mytexttt{GE-001}: \mytexttt{gemini-embedding-001} \\

\textbf{Benchmarks:} COIR-CSN: COIRCodeSearchNetRretrieval; CSN: CodeSearchNetRetrieval; CES: CodeEditSearchRetrieval; CP-FT: CommitPackFT; AppsR: AppsRetrieval; SynText2SQL: SyntheticText2SQL: CF-MT: CodeFeedbackMT; CF-ST: CodeFeedbackST; StackOQA: StackOverflowQA; CTOC: CodeTransOceanContest; CTODL: CodeTransOceanDL; CSN-CC: CodeSearchNetCCRetrieval\\
* Benchmarks of the MTEB Code leaderboard.\\
}
\end{table}

Both \modelS{} and \modelLshort{} outperform similar-sized general-purpose embedding model \texttt{Qwen3-Embedding-0.6B} and the substantially larger models \texttt{jina-embeddings-v4} and \texttt{gemini-embedding-001}.

\section{Conclusion}

We have introduced \model{}, a family of code embedding models with 0.5B and 1.5B parameters. By using an autoregressive backbone pre-trained on both text and code, along with task-specific instruction prefixes and last-token pooling, the models excel at a wide variety of tasks and domains related to code retrieval. Despite their smaller size compared to other models, the \model{} suite achieves state-of-the-art performance, demonstrating the validity and effectiveness of its unique construction methodology.



\newpage

\bibliographystyle{plainnat}
\bibliography{custom}

\newpage

\appendix

\section{Training Datasets}\label{datasets}

Training data for \model{} draws on a variety of sources, described in Section~\ref{sec:training-data}. 

\begin{itemize}
    \item Training data splits for MTEB code tasks, and the CoSQA+ dataset.
    \item Other public datasets adapted to our training needs.
    \item Fully or partially synthetic datasets generated using GPT-4o ~\citep{openai-2024}.
\end{itemize}

The \textbf{SyntheticDLTrans} dataset consists of generated deep learning code translations between frameworks. Since very little non-synthetic data of this kind is available, we generated all query-document pairs from scratch by repeatedly prompting GPT-4o to output a pair of equivalent deep learning code snippets in two specified distinct frameworks (e.g., PyTorch and TensorFlow). We instructed GPT-4o that the code snippets can define model architectures (Transformers, CNNs, RNNs, simple deep networks, etc.), training loops, evaluations, and mathematical operations on tensors.

We also synthesized a multilingual extension of the CodeChef dataset~\citep{caballero-2016}, using the original programming solutions in C++ and Python to generate solutions in eight more programming languages (C, C\#, Go, Java, JavaScript, Ruby, Rust, TypeScript). We prompted GPT-4o to produce an equivalent solution in a specified new language given one C++ solution and one Python solution as references (but we did not use problem statements for generation). The resulting dataset has been adapted for three tasks: \textbf{CodeChefP2S} (problem-to-solution), \textbf{CodeChefS2S} (monolingual solution-to-solution), and \textbf{CodeChefXLang} (crosslingual solution-to-solution).

Synthetic examples were validated by manual inspection of samples.

Table~\ref{tab:training-datasets} provides details about the provenance of all training datasets. Table~\ref{tab:training-cats} shows the task categories of each training dataset, and Table~\ref{tab:eval-cats} does the same for the evaluation datasets.

\begin{table}[h!]
\caption{Datasets used to train \model{}}
\label{tab:training-datasets}
\scriptsize
\begin{tabular}{l|l|l}
\textbf{Dataset} & \textbf{Type} &\textbf{Source} \\
\hline
AppsRetrieval & \multirow{10}{*}{MTEB Code} & \url{https://huggingface.co/datasets/CoIR-Retrieval/apps} \\
CodeFeedbackMT & & \url{https://huggingface.co/datasets/CoIR-Retrieval/codefeedback-mt} \\
CodeFeedbackST & & \url{https://huggingface.co/datasets/CoIR-Retrieval/codefeedback-st} \\
CodeTransOceanContest & & \url{https://huggingface.co/datasets/CoIR-Retrieval/codetrans-contest} \\
CodeTransOceanDL & & \url{https://huggingface.co/datasets/CoIR-Retrieval/codetrans-dl} \\
CodeSearchNetCCRetrieval & & \url{https://huggingface.co/datasets/CoIR-Retrieval/CodeSearchNet-ccr} \\
COIR-CodeSearchNet & & \url{https://huggingface.co/datasets/CoIR-Retrieval/CodeSearchNet} \\
CoSQA & & \url{https://huggingface.co/datasets/CoIR-Retrieval/cosqa} \\
StackOverflowQA & & \url{https://huggingface.co/datasets/CoIR-Retrieval/stackoverflow-qa} \\
SyntheticText2SQL & & \url{https://huggingface.co/datasets/CoIR-Retrieval/synthetic-text2sql} \\
\hline
CodeForcesP2S & \multirow{19}{*}{Adapted} & \url{https://huggingface.co/datasets/MatrixStudio/Codeforces-Python-Submissions} \\
CodeForcesS2S & & \url{https://github.com/ethancaballero/description2code} \\
CodeSearchNet & & \url{https://github.com/github/CodeSearchNet} \\
CommitPackFT & & \url{https://huggingface.co/datasets/bigcode/commitpackft} \\
CoSQA+ & & \url{https://github.com/DeepSoftwareAnalytics/CoSQA_Plus} \\
DataScience & & \url{https://kaggle.com/datasets/stackoverflow/stacksample} \\
Doc2Code & & \url{https://github.com/EdinburghNLP/code-docstring-corpus} \\
GlaiveCodeAssistantV2 & & \url{https://huggingface.co/datasets/glaiveai/glaive-code-assistant-v2} \\
HackerEarth & & \url{https://github.com/ethancaballero/description2code} \\
LeetCodeP2S & & \multirow{2}{*}{\url{https://huggingface.co/datasets/greengerong/leetcode}} \\
LeetCodeXLang & & \\
MBPP & & \url{https://huggingface.co/datasets/google-research-datasets/mbpp} \\
MLQuestions & & \url{https://huggingface.co/datasets/McGill-NLP/mlquestions} \\
Spider & & \url{https://huggingface.co/datasets/xlangai/spider} \\
StackExchangeBody & & \multirow{3}{*}{\url{https://github.com/EleutherAI/stackexchange_dataset/}} \\
StackExchangePost & & \\
StackExchangeTitle & & \\
SWE-Bench & & \url{https://huggingface.co/datasets/princeton-nlp/SWE-bench} \\
WikiSQL & & \url{https://huggingface.co/datasets/Salesforce/wikisql} \\
\hline
CodeChefP2S & \multirow{4}{*}{Synthetic} & \\
CodeChefS2S & & \\
CodeChefXLang & & \\
SyntheticDLTrans & & \\
\hline
\end{tabular}
\vspace{3mm}
\end{table}

\begin{table}[h!]
\caption{Breakdown of the training (a) and evaluation (b) datasets by task type.}
\label{tab:tasks}
\scriptsize
    \begin{subtable}[c]{0.5\textwidth}
    \subcaption{Training datasets}
    \label{tab:training-cats}
    \centering
    \begin{tabular}{l|c}
    \toprule
    \textbf{Dataset} & \textbf{Task type} \\
    \hline
    AppsRetrieval & \multirow{14}{*}{NL2Code} \\
    CodeChefP2S & \\
    CodeForcesP2S & \\
    CodeSearchNet & \\
    CommitPackFT & \\
    CoSQA & \\
    CoSQA+ & \\
    Doc2Code & \\
    LeetCodeP2S & \\
    MBPP & \\
    Spider & \\
    SWE-Bench & \\
    SyntheticText2SQL & \\
    WikiSQL & \\
    \hline
    CodeFeedbackMT & \multirow{9}{*}{TechQA} \\
    CodeFeedbackST & \\
    DataScience & \\
    GlaiveCodeAssistantV2 & \\
    MLQuestions & \\
    StackExchangeBody & \\
    StackExchangePost & \\
    StackExchangeTitle & \\
    StackOverflowQA & \\
    \hline
    CodeChefS2S & \multirow{8}{*}{Code2Code} \\
    CodeChefXLang & \\
    CodeForcesS2S & \\
    CodeTransOceanContest & \\
    CodeTransOceanDL & \\
    HackerEarth & \\
    LeetCodeXLang & \\
    SyntheticDLTrans & \\
    \hline
    COIRCodeSearchNetRetrieval & Code2NL \\
    \hline
    CodeSearchNetCCRetrieval & Code2Completion \\
    \bottomrule
    \end{tabular}
    \end{subtable}
    \begin{subtable}[c]{0.5\textwidth}
    \subcaption{Evaluation datasets}
    \label{tab:eval-cats}
    \centering
    \begin{tabular}{l|c}
    \toprule
    \textbf{Dataset} & \textbf{Task type} \\
    \hline
    AppsRetrieval & \multirow{15}{*}{NL2Code} \\
    CodeChef & \\
    CodeEditSearchRetrieval & \\
    CodeSearchNetRetrieval & \\
    CommitPackFT & \\
    CoSQA & \\
    CoSQA+ & \\
    Doc2Code & \\
    HumanEval & \\
    LeetCode & \\
    MBPP & \\
    Spider & \\
    SWE-Bench & \\
    SyntheticText2SQL & \\
    WikiSQL & \\
    \hline
    CodeFeedbackMT & \multirow{5}{*}{TechQA} \\
    CodeFeedbackST & \\
    DS-1000 & \\
    MLQuestions & \\
    StackOverflowQA & \\
    \hline
    CodeChefXLang & \multirow{3}{*}{Code2Code} \\
    CodeTransOceanContest & \\
    CodeTransOceanDL & \\
    \hline
    COIRCodeSearchNetRetrieval & Code2NL \\
    \hline
    CodeSearchNetCCRetrieval & Code2Completion \\
    \bottomrule
    \end{tabular}
    \end{subtable}
\end{table}

\clearpage
\newpage

\section{Ablation}\label{ablation}

We trained three versions of the 0.5B model with the same training data, hyperparameters, and number of steps (1500), but with three different pooling methods: last-token, mean, and latent-attention. We found that last-token pooling results in the highest average performance (see Table \ref{tab:ablation}).

\begin{table}[h]
\small
\centering
\caption{Results of the pooling ablation experiments.}
\begin{tabular}{l|c|c|c}
\toprule
\textbf{Benchmark} & \textbf{Last-token} & \textbf{Mean} & \textbf{Latent attention} \\
\hline
CoSQA+ & 15.42\% & 15.36\% & \textbf{15.55\%} \\
CoSQA* & \textbf{39.25\%} & 37.13\% & 38.58\% \\
MBPP & \textbf{89.01\%} & 87.01\% & 88.57\% \\
COIR-CSN* & \textbf{85.73\%} & 85.01\% & 85.50\% \\
CSN* & \textbf{90.68\%} & 90.48\% & 90.65\% \\
Doc2Code & \textbf{95.98\%} & 95.91\% & 95.94\% \\
SWE-Bench & 83.00\% & \textbf{83.88\%} & 83.31\% \\
CES* & \textbf{83.25\%} & 82.94\% & 83.09\% \\
CP-FT & 63.00\% & 62.32\% & \textbf{63.10\%} \\
AppsR* & 84.17\% & 83.26\% & \textbf{84.43\%} \\
LeetCode & 57.86\% & 58.08\% & \textbf{58.17\%} \\
CodeChef & 94.03\% & 92.08\% & \textbf{95.03\%} \\
SynText2SQL* & 72.80\% & 72.60\% & \textbf{72.93\%} \\
Spider & 81.65\% & \textbf{81.99\%} & 81.57\% \\
WikiSQL & \textbf{98.31\%} & 93.50\% & 97.85\% \\
CF-MT* & \textbf{89.56\%} & 86.09\% & 88.95\% \\
CF-ST* & \textbf{85.73\%} & 84.55\% & 85.23\% \\
StackOQA* & \textbf{91.04\%} & 90.46\% & 90.58\% \\
DS-1000 & 59.77\% & 58.91\% & \textbf{60.20\%} \\
MLQuestions & 81.05\% & 79.78\% & \textbf{81.07\%} \\
CTOC* & 90.37\% & 86.85\% & \textbf{90.70\%} \\
CTODL* & \textbf{41.69\%} & 38.17\% & 40.58\% \\
CodeChefXLang & \textbf{99.70\%} & 99.31\% & 99.21\% \\
CSN-CC* & \textbf{90.41\%} & 88.65\% & 89.72\% \\
HumanEval* & \textbf{96.77\%} & 95.78\% & 96.35\% \\
\hline
Overall AVG & \textbf{78.41\%} & 77.20\% & 78.27\% \\
MTEB Code AVG & \textbf{78.72\%} & 77.18\% & 78.41\% \\
\bottomrule
\end{tabular}
\label{tab:ablation}
\end{table}

\section{Evaluation Hyperparameters}\label{sec:eval-hparams}

\model{} and \mytexttt{Qwen3-Embedding-0.6B} were evaluated in FP16 with a batch size of 8 and a sequence length of 8192; \mytexttt{jina-embeddings-v4} was evaluated in BF16 with a batch size of 8 and a sequence length of 8192. The Voyage and Gemini models were evaluated via the respective APIs with a batch size of 8, except the tasks COIRCodeSearchNetRetrieval and CodeSearchNetCCRetrieval, which we did not evaluate due to the large size of the benchmarks and the resulting cost in time and money, so we took public scores from the MTEB GitHub.

\end{document}